\documentclass{article} 
\usepackage{iclr2024_conference,times}

\usepackage{amsmath,amsfonts,bm}

\def\eqref#1{equation~\ref{#1}}

\def\1{\bm{1}}

\DeclareMathAlphabet{\mathsfit}{\encodingdefault}{\sfdefault}{m}{sl}
\SetMathAlphabet{\mathsfit}{bold}{\encodingdefault}{\sfdefault}{bx}{n}

\usepackage{multirow}
\usepackage{xcolor}
\usepackage{array}
\usepackage{graphicx}
\usepackage{hyperref}
\usepackage{url}
\usepackage{booktabs}

\usepackage{pifont} 
\newcommand{\cmark}{\ding{51}}%
\newcommand{\xmark}{\ding{55}}%

\title{Red Teaming Models for Hyperspectral Image Analysis Using Explainable AI}

\author{
Vladimir Zaigrajew\textsuperscript{$1$} \And Hubert Baniecki\textsuperscript{$1$,$2$} \And  
Lukasz Tulczyjew\textsuperscript{$3$,$4$} \And Agata M. Wijata\textsuperscript{$3$,$4$} \And Jakub Nalepa\textsuperscript{$3$,$4$} \And Nicolas Longépé\textsuperscript{$5$} \And
Przemyslaw Biecek\textsuperscript{$1$,$2$} \and\\
\textsuperscript{$1$}Warsaw University of Technology,
\textsuperscript{$2$}University of Warsaw,
\textsuperscript{$3$}Silesian University of Technology,\\
\textsuperscript{$4$}KP Labs,
\textsuperscript{$5$}European Space Agency\\
\textsuperscript{$1$}\texttt{vladimir.zaigrajew.dokt@pw.edu.pl}\\
}

\newcommand{\Hyperview}{\textsc{hyperview}\,}
\newcommand{\Intuition}{\textsc{intuition-1}\,}
\newcommand{\EagleEyes}{\textsc{eagleeyes}\,}
\definecolor{darkgreenmy}{rgb}{0,0.5,0.3}

\iclrfinalcopy 
\begin{document}

\maketitle

\begin{abstract}
Remote sensing (RS) applications in the space domain demand machine learning (ML) models that are reliable, robust, and quality-assured, making red teaming a vital approach for identifying and exposing potential flaws and biases. Since both fields advance independently, there is a notable gap in integrating red teaming strategies into RS. This paper introduces a methodology for examining ML models operating on hyperspectral images within the \Hyperview challenge, focusing on soil parameters' estimation. We use post-hoc explanation methods from the Explainable AI (XAI) domain to critically assess the best performing model that won the \Hyperview challenge and served as an inspiration for the model deployed on board the \Intuition hyperspectral mission. Our approach effectively red teams the model by pinpointing and validating key shortcomings, constructing a model that achieves comparable performance using just $1\%$ of the input features and a mere up to $5\%$ performance loss. Additionally,  we propose a novel way of visualizing explanations that integrate domain-specific information about hyperspectral bands (wavelengths) and data transformations to better suit interpreting models for hyperspectral image analysis.
\end{abstract}

\section{Introduction}

Remotely-sensed hyperspectral imaging presents significant opportunities in Earth observation, offering an extensive amount of data and potential for global scalability~\citep{wijata2023taking}. Hyperspectral images (HSIs) capture numerous spectral bands, providing detailed insights about scanned objects. In precision agriculture, HSIs are crucial for extracting various soil parameters, enabling non-invasive, large-scale monitoring to enhance agricultural practices~\citep{nalepa2022hyperview}. However, deploying machine learning (ML) models for these tasks faces challenges in creating diverse, high-quality datasets, often constrained by time, cost, and human error. The \Hyperview: ``Seeing beyond the visible'' challenge, with its carefully curated dataset, aimed to address these issues by developing models to retrieve soil parameters from HSIs. Attracting substantial participation (160 registered teams, with ca. 50 actively participating) from the research community~\citep{nalepa2022hyperview}, the challenge resulted in a high-performing ML model. The winning solution was set for deployment on the \Intuition satellite. However, before its deployment on the \Intuition satellite, the winning challenge solution had to undergo a thorough performance evaluation. 

This important deployment phase reflects the wider industry trend of continuous model improvement, where the use of various \emph{red teaming} strategies is essential. These strategies aim to uncover flaws and biases across diverse model families such as large language models~\citep{perez2022red,ganguli2022red}, text-to-image models~\citep{mehrabi2023flirt}, and diffusion models~\citep{rando2022redteaming}. In parallel, stakeholders in remote sensing (RS) call for higher explainability standards~\citep{roscher2020explain,gevaert2022explainable} as explainable AI (XAI) methods have shown to be useful in many other practical applications~\citep{roscher2020explainable}. Therefore, several works~\citep{singh2022estimation, turan2023interpreting, dantas2023counterfactual, emam2023leveraging, ekim2023explaining, abbas2023towards} have emerged to explain models used in RS. However, only a few studies have shown how applying XAI can pinpoint performance gaps in models~\citep{emam2023confident} or improve predictions~\citep{de2022towards}, effectively aligning XAI with \emph{red teaming} strategies. Our work aims to advance this integration, focusing specifically on models from the \Hyperview challenge designed for estimating soil parameters. We adapt post-hoc explanations to the specific domain and use existing knowledge about the model to generate more sophisticated explanatory visualizations, which are used to successfully red team the model.

\paragraph{Contribution.}
Our key contributions are: (1) We demonstrate how, by using SHapley Additive exPlanations (SHAP), we were able to conduct an in-depth analysis of the \EagleEyes model~\citep{kuzu2022predicting} within the \Hyperview challenge, laying the groundwork for unraveling shortcomings in model performance. (2) We present a novel way of visualizing explanations that integrate domain-specific information about hyperspectral bands and data transformations, aimed at boosting the red teaming of hyperspectral image analysis models. (3) The result, aimed at red teaming the model, was the development of a model pruning technique focused on feature selection. This method, based on SHAP, yielded a more efficient model without compromising its regression performance.

\section{Methodology}

\subsection{Shapley values as a model agnostic tool for  knowledge extraction}

Shapley values for predictive models~\citep{robnik2008explaining,lundberg2017unified} are one of the most popular XAI methods~\citep{Holzinger2022}. This well-deserved popularity is due to such properties as model agnosticity (they can be calculated for any model structure), additivity (so that the contribution of a group of features is the sum of the contributions of the features in that group), the availability of efficient implementations (e.g., TreeSHAP for tree-based models~\citep{lundberg2020local}), and the possibility of use in the process of model analysis~\citep{ema2021}. For a detailed mathematical explanation, please refer to Appendix~\ref{sec:math}. Shapley values are initially computed locally for individual features and observations, but they can also be aggregated to get the global importance of a selected variable, or attribution of a group of features.
In the following sections, we will aggregate features corresponding to the same spectral band or the same data transformation method.

\subsection{Data And Models}\label{sec:challenge}

We work with the \Hyperview dataset and the winning \Hyperview model \EagleEyes. The dataset, acquired on March 3, 2021, using the HySpex VS-725 (Norsk Elektro Optikk AS) acquisition system, consists of images from two SWIR-384 imagers and one VNIR-1800 imager aboard the Piper PA-31 Navajo aircraft. After undergoing comprehensive preprocessing and calibration as detailed by~\citet{nalepa2022hyperview}, the images were aligned with in-situ soil parameter measurements taken using the Mehlich 3 methodology. From each parcel, 12 soil samples were collected and analyzed in the laboratory. This process provided single, weakly-labeled, per-image ground-truth data for four key soil parameters: \textit{phosphorus} (\textit{P}), \textit{potassium}, (\textit{K}) \textit{magnesium} (\textit{Mg}) and soil acidity ({\textit{pH}}). Our study analyzes two dataset versions: \Hyperview with 150 bands (462--942\,nm) and \Intuition with 192 bands (460--920\,nm), matching the spectral characteristics of a hyperspectral camera mounted on board Intuition-1, a 6U satellite (KP Labs, Poland). The \EagleEyes model, available at \url{https://github.com/ridvansalihkuzu/hyperview_eagleeyes}, was provided with and without spatial features extracted and deployed over two aforementioned versions of the dataset. For detailed information on the dataset and model, refer to Appendix~\ref{sec:data_anal}.

\subsection{Model Pruning}\label{sec:pruning}

ML models, particularly complex ones, require substantial computational resources for training and inference. This is challenging for real-time processing and deployment on resource-constrained devices like imaging satellites. Model pruning, a technique to reduce model size without affecting performance, addresses this issue. It is widely used in both classic ML~\citep{zhou2023trees} and deep learning~\citep{cheng2023survey} to optimize models for specific tasks, including those in RS~\citep{qi2019network, guo2021network, zhou2018feature}. However, traditional pruning methods often depend on time-consuming processes and can introduce constraints. Our approach utilizes explanations derived from Shapley values for effective feature selection in model pruning. This method significantly reduces the model's size and computational demands while preserving its performance, resulting in lighter and faster models suitable for edge devices.

\section{Explanatory Model Analysis}

In this section, we conduct a thorough analysis of the \Hyperview challenge, focusing on evaluating the top-ranked model's performance through Shapley values. Our goal is to identify and highlight any shortcomings in these models, offering insights for enhancements in their ongoing deployment. The initial analysis of the models and datasets is detailed in Appendix~\ref{sec:data_anal}. A critical finding was the poor model performance, notably in consistently predicting $90\%$ of values within a narrow range, as seen in Figure~\ref{fig:residuals_all}. This led to significant mispredictions, especially for outliers in each soil parameter.

\paragraph{SHAP}
Our initial Shapley values analysis examined feature importance distribution across soil parameters and model versions, with detailed visualizations in Appendix~\ref{sec:shap_vis}. Figure~\ref{fig:shap_all_soil} displays the top features of the \EagleEyes model and their importance in predicting each soil parameter. Meanwhile, Figure~\ref{fig:shap_all_models} provides beeswarm plots for each version of the \EagleEyes model in predicting \textit{phosphorus}, highlighting the top features and their impact on predictions as defined by Shapley values. Key findings include the models relying on a limited set of features (less than $1\%$ of the total), likely causing their narrow prediction range. Each model version favors different features, and as the number of features used increases, the focus shifts from the most influential ones, hinting at possible \textbf{underfitting} and reinforcing the limited prediction range observed.

\begin{figure}[h]
\centering
\includegraphics[width=1.0\textwidth]{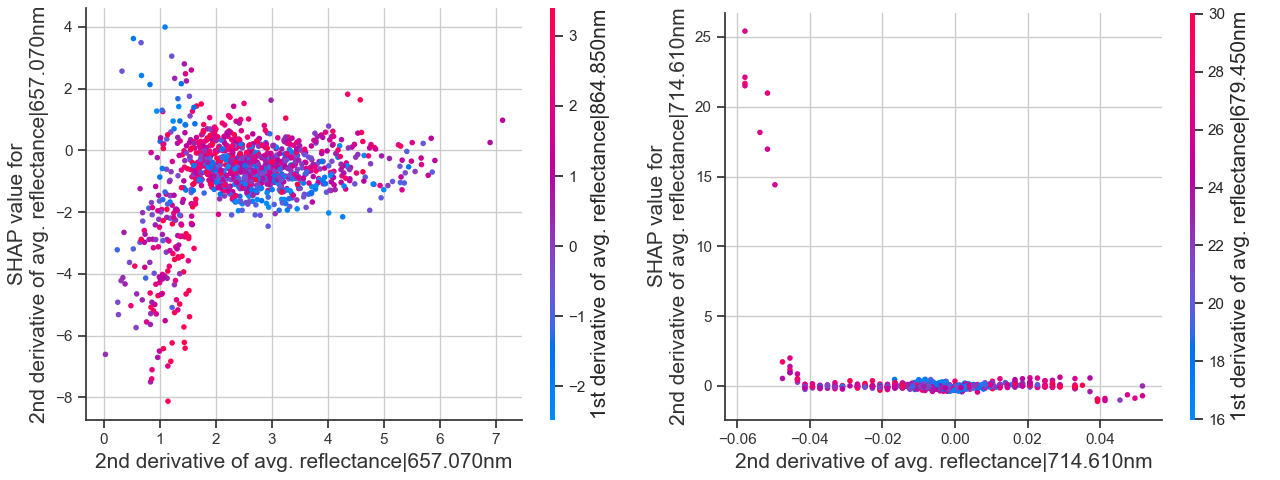}
\caption{Variations in Shapley values in response to changing feature values in the \EagleEyes model for \textit{phosphorus} predictions, highlighting the most influential feature (\textbf{left}) and a feature with lesser impact (\textbf{right}).}
\label{fig:shap_hyperview}
\end{figure}

While popular feature importance methods provide similar insights, Shapley values offer a more detailed analysis of how individual features and specific samples impact model predictions. Analyzing SHAP's dependency and beeswarm plots, we gain an understanding of the effects of individual feature value fluctuations on predictions. Figure~\ref{fig:shap_hyperview} illustrates this in two parts: the left plot demonstrates how Shapley values correlate with the most impactful feature in the \EagleEyes model, peaking at a value of $2$ and then stabilizing, suggesting a limited utilization of feature values for even most impactful features. The right plot, targeting a less influential feature, indicates the model's bias towards memorizing specific outliers, resulting in prediction variability around the established prediction window. These observations support our hypothesis regarding the model's constrained prediction range, which primarily depends on key features for \textbf{simplistic} value estimation and incorporates \textbf{outlier memorization} for adaptability within its prediction range.

\begin{figure}[h]
\centering
\includegraphics[width=0.99\textwidth]{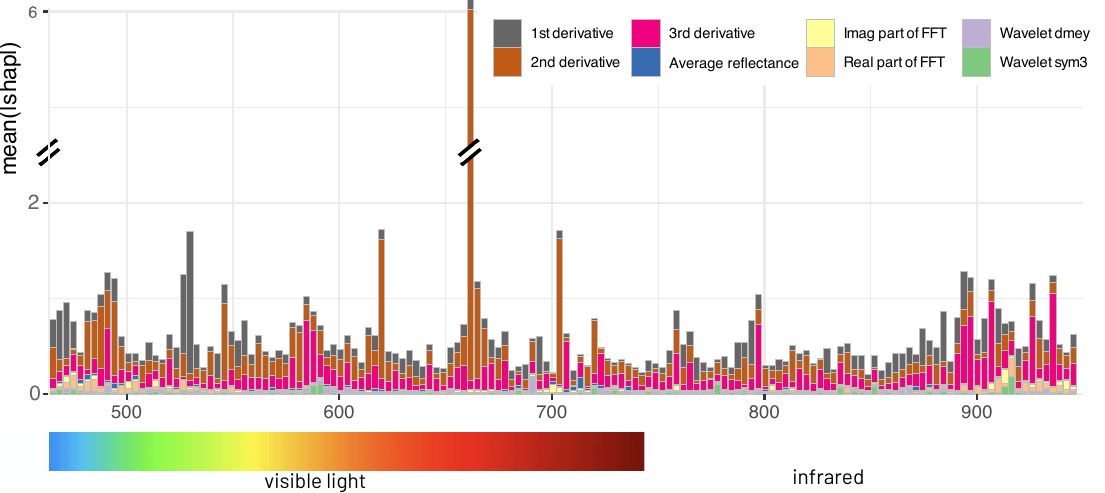}
\caption{Detailed visualization of Shapley values, depicting the significance of various aggregated feature transformation groups with their corresponding wavelengths aggregations on the $x$-axis.}
\label{fig:shap_aggregation}
\end{figure}

\paragraph{Aggregation Analysis}
Shapley values, formatted as \texttt{<n\_samples, features, class>}, enable aggregation into \textbf{hyperspectral bands} and \textbf{data transformation} groups. Figure~\ref{fig:shap_aggregation} demonstrates the significance of features, grouped by preprocessing transformations on the $y$-axis and wavelength (hyperspectral band) on the $x$-axis. This visualization aids in understanding model interpretations within the spectral domain of HSIs, incorporating preprocessing transformations. Notably, this plot reveals that key features are distributed across various bands instead of being clustered in specific areas. Aggregated transformation visualizations, detailed in Appendix~\ref{sec:aggr_vis}, offer more informative insights. Figure~\ref{fig:shap_transformation} highlights that spectral gradient features significantly influence the \EagleEyes model's predictions, unlike other transformations with minimal impact. Additionally, models incorporating spatial data tend to rely more on spatial features for predictions, underscoring their critical role in enhancing model performance.

\paragraph{Model Pruning}

In our concluding analysis, we assessed model performance using key features identified through Shapley value analysis. Beeswarm and feature importance plots enabled the efficient selection of a feature subset for the prediction of soil parameters. Table~\ref{table:feature_selection} shows that models with a reduced number of input features reached Mean Absolute Error (MAE) scores comparable to those using all features for each soil parameter. Models trained with spatial features presented challenges in selecting feature subsets for training. It was expected due to smoother feature importance distribution. Nonetheless, pruned models with under $1\%$ of features performed similarly to baseline models, as a two-sample t-test confirmed, with minimal metric decline. This affirms our hypothesis about the models' dependency on a limited set of features for precise predictions.

\begin{table}[h]
\centering
\small
\begin{tabular}{ ll|llll } 
 \toprule
 \textbf{Dataset} & \textbf{F. selection} & \textit{P} & \textit{K} & \textit{Mg} & \textit{pH} \\ 
 \midrule
 \multirow{2}{*}{\Hyperview} 
 & \xmark\;(1200) & 22.6 & 48.4 & 31.3 & 0.206 \\ 
 & \cmark\;(3) & 23.3 ($+3\%$) & 48.7 ($+1\%$) & 32.6 ($+4\%$) & 0.213 ($+4\%$) \\
 \midrule
 \multirow{2}{*}{\Intuition} 
 & \xmark\;(1490) & 22.5 & 48.5 & 30.9 & 0.204 \\
 & \cmark\;(3) & 22.7 ($+1\%$) & 48.5 ($0\%$) & 32.5 ($+5\%$) &  0.214 ($+5\%$) \\
 \midrule
 \multirow{2}{*}{\Hyperview (spatial)} 
 & \xmark\;(2400) & 22.4 & 47.8 & 31.0 & 0.205 \\
 & \cmark\;(5) & 23.0 ($+3\%$) & 48.9 ($+2\%$) & 32.5 ($+5\%$) & 0.212 ($+3\%$) \\
 \midrule
 \multirow{2}{*}{\Intuition (spatial)} 
 & \xmark\;(3026) & 22.3 & 47.8 & 31.1 & 0.204 \\
 & \cmark\;(6) & 23.2 ($+4\%$) & 50.1 ($+5\%$) & 32.4 ($+4\%$) & 0.209 ($+2\%$) \\
 \bottomrule
\end{tabular}
\caption{Comparison of model performance on the test dataset using MAE ($\downarrow$) for soil parameters (\textit{P}, \textit{K}, \textit{Mg}, \textit{pH}). Evaluations cover two scenarios: trained on original features (\xmark) and trained on post-feature selection (\cmark), noting the feature count, e.g., 3 of 1200 for \Hyperview.}
\label{table:feature_selection}
\end{table}



\section{Conclusion}

We developed an innovative red teaming methodology for hyperspectral image analysis models, focusing on SHAP to utilize explainability and apply assurance of robustness. Our method, employed on the winning \Hyperview challenge model (\EagleEyes), uncovers and visualizes model failures and biases, offering advanced visualizations into hyperspectral bands and data transformation aggregations. Notably, we found the \EagleEyes model utilized less than $1\%$ of available features, leading us to create simpler but effective model alternatives. Our study marks a significant advance in the application of red teaming using explainable AI to the hyperspectral imaging domain models.

\section*{Acknowledgements} We thank Bogdan Ruszczak for his valuable feedback about this work. The work on this project is financially supported by the European Space Agency grant (ESA AO/1-11524/22/I-DT) and it was supported by $\Phi$-lab (https://philab.esa.int/). Lukasz Tulczyjew, Agata M. Wijata and Jakub Nalepa were supported by the Silesian University of Technology through the grant for maintaining and developing research potential.

\bibliography{references}

\begin{thebibliography}{29}
\providecommand{\natexlab}[1]{#1}
\providecommand{\url}[1]{\texttt{#1}}
\expandafter\ifx\csname urlstyle\endcsname\relax
  \providecommand{\doi}[1]{doi: #1}\else
  \providecommand{\doi}{doi: \begingroup \urlstyle{rm}\Url}\fi

\bibitem[Abbas et~al.(2023)Abbas, Linardi, Vareille, Christophides, and Paris]{abbas2023towards}
Adel Abbas, Michele Linardi, Etienne Vareille, Vassillis Christophides, and Claudia Paris.
\newblock Towards explainable ai4eo: An explainable deep learning approach for crop type mapping using satellite images time series.
\newblock In \emph{IGARSS 2023-2023 IEEE International Geoscience and Remote Sensing Symposium}, pp.\  1088--1091. IEEE, 2023.

\bibitem[Biecek \& Burzykowski(2021)Biecek and Burzykowski]{ema2021}
Przemyslaw Biecek and Tomasz Burzykowski.
\newblock \emph{{Explanatory Model Analysis}}.
\newblock Chapman and Hall/CRC, New York, 2021.

\bibitem[Cheng et~al.(2023)Cheng, Zhang, and Shi]{cheng2023survey}
Hongrong Cheng, Miao Zhang, and Javen~Qinfeng Shi.
\newblock A survey on deep neural network pruning-taxonomy, comparison, analysis, and recommendations.
\newblock \emph{arXiv preprint arXiv:2308.06767}, 2023.

\bibitem[Covert et~al.(2020)Covert, Lundberg, and Lee]{covert2020understanding}
Ian Covert, Scott~M Lundberg, and Su-In Lee.
\newblock {Understanding Global Feature Contributions With Additive Importance Measures}.
\newblock In \emph{NeurIPS}, 2020.

\bibitem[Dantas et~al.(2023)Dantas, Drumond, Marcos, and Ienco]{dantas2023counterfactual}
Cassio~F. Dantas, Thalita~F. Drumond, Diego Marcos, and Dino Ienco.
\newblock {Counterfactual Explanations for Remote Sensing Time Series Data: An Application to Land Cover Classification}.
\newblock In \emph{ECML PKDD}, 2023.

\bibitem[De~Lucia et~al.(2022)De~Lucia, Lapegna, and Romano]{de2022towards}
Gianluca De~Lucia, Marco Lapegna, and Diego Romano.
\newblock Towards explainable ai for hyperspectral image classification in edge computing environments.
\newblock \emph{Computers and Electrical Engineering}, 103:\penalty0 108381, 2022.

\bibitem[Ekim \& Schmitt(2023)Ekim and Schmitt]{ekim2023explaining}
Burak Ekim and Michael Schmitt.
\newblock {Explaining Multimodal Data Fusion: Occlusion Analysis for Wilderness Mapping}.
\newblock In \emph{ICLR ML4RS Workshop}, 2023.

\bibitem[Emam et~al.(2023{\natexlab{a}})Emam, Roscher, and Farag]{emam2023confident}
Ahmed Emam, Ribana Roscher, and Mohamed Farag.
\newblock Confident naturalness explanation (cne): A framework to explain and assess patterns forming naturalness in fennoscandia with confidence.
\newblock In \emph{Northern Lights Deep Learning Conference}, number FZJ-2024-00529. Pflanzenwissenschaften, 2023{\natexlab{a}}.

\bibitem[Emam et~al.(2023{\natexlab{b}})Emam, Stomberg, and Roscher]{emam2023leveraging}
Ahmed Emam, Timo~T Stomberg, and Ribana Roscher.
\newblock Leveraging activation maximization and generative adversarial training to recognize and explain patterns in natural areas in satellite imagery.
\newblock \emph{IEEE Geoscience and Remote Sensing Letters}, 2023{\natexlab{b}}.

\bibitem[Ganguli et~al.(2022)Ganguli, Lovitt, Kernion, Askell, Bai, Kadavath, Mann, Perez, Schiefer, Ndousse, et~al.]{ganguli2022red}
Deep Ganguli, Liane Lovitt, Jackson Kernion, Amanda Askell, Yuntao Bai, Saurav Kadavath, Ben Mann, Ethan Perez, Nicholas Schiefer, Kamal Ndousse, et~al.
\newblock Red teaming language models to reduce harms: Methods, scaling behaviors, and lessons learned.
\newblock \emph{arXiv preprint arXiv:2209.07858}, 2022.

\bibitem[Gevaert(2022)]{gevaert2022explainable}
Caroline~M Gevaert.
\newblock Explainable ai for earth observation: A review including societal and regulatory perspectives.
\newblock \emph{International Journal of Applied Earth Observation and Geoinformation}, 112:\penalty0 102869, 2022.

\bibitem[Guo et~al.(2021)Guo, Hou, Ren, Ren, and Jiao]{guo2021network}
Xianpeng Guo, Biao Hou, Bo~Ren, Zhongle Ren, and Licheng Jiao.
\newblock Network pruning for remote sensing images classification based on interpretable cnns.
\newblock \emph{IEEE Transactions on Geoscience and Remote Sensing}, 60:\penalty0 1--15, 2021.

\bibitem[Holzinger et~al.(2022)Holzinger, Saranti, Molnar, Biecek, and Samek]{Holzinger2022}
Andreas Holzinger, Anna Saranti, Christoph Molnar, Przemyslaw Biecek, and Wojciech Samek.
\newblock {Explainable AI Methods -- A Brief Overview}.
\newblock In \emph{xxAI - Beyond Explainable AI: International Workshop, Held in Conjunction with ICML 2020}, pp.\  13--38, 2022.

\bibitem[Kuzu et~al.(2022)Kuzu, Albrecht, Arnold, Kamath, and Konen]{kuzu2022predicting}
R{\i}dvan~Salih Kuzu, Frauke Albrecht, Caroline Arnold, Roshni Kamath, and Kai Konen.
\newblock Predicting soil properties from hyperspectral satellite images.
\newblock In \emph{2022 IEEE International Conference on Image Processing (ICIP)}, pp.\  4296--4300. IEEE, 2022.

\bibitem[Lundberg \& Lee(2017)Lundberg and Lee]{lundberg2017unified}
Scott~M. Lundberg and Su-In Lee.
\newblock {A Unified Approach to Interpreting Model Predictions}.
\newblock In \emph{NeurIPS}, 2017.

\bibitem[Lundberg et~al.(2020)Lundberg, Erion, Chen, DeGrave, Prutkin, Nair, Katz, Himmelfarb, Bansal, and Lee]{lundberg2020local}
Scott~M Lundberg, Gabriel Erion, Hugh Chen, Alex DeGrave, Jordan~M Prutkin, Bala Nair, Ronit Katz, Jonathan Himmelfarb, Nisha Bansal, and Su-In Lee.
\newblock From local explanations to global understanding with explainable ai for trees.
\newblock \emph{Nature machine intelligence}, 2\penalty0 (1):\penalty0 56--67, 2020.

\bibitem[Mehrabi et~al.(2023)Mehrabi, Goyal, Dupuy, Hu, Ghosh, Zemel, Chang, Galstyan, and Gupta]{mehrabi2023flirt}
Ninareh Mehrabi, Palash Goyal, Christophe Dupuy, Qian Hu, Shalini Ghosh, Richard Zemel, Kai-Wei Chang, Aram Galstyan, and Rahul Gupta.
\newblock Flirt: Feedback loop in-context red teaming.
\newblock \emph{arXiv preprint arXiv:2308.04265}, 2023.

\bibitem[Nalepa et~al.(2022)Nalepa, Le~Saux, Long{\'e}p{\'e}, Tulczyjew, Myller, Kawulok, Smykala, and Gumiela]{nalepa2022hyperview}
Jakub Nalepa, Bertrand Le~Saux, Nicolas Long{\'e}p{\'e}, Lukasz Tulczyjew, Michal Myller, Michal Kawulok, Krzysztof Smykala, and Michal Gumiela.
\newblock The hyperview challenge: Estimating soil parameters from hyperspectral images.
\newblock In \emph{2022 IEEE International Conference on Image Processing (ICIP)}, pp.\  4268--4272. IEEE, 2022.

\bibitem[Perez et~al.(2022)Perez, Huang, Song, Cai, Ring, Aslanides, Glaese, McAleese, and Irving]{perez2022red}
Ethan Perez, Saffron Huang, Francis Song, Trevor Cai, Roman Ring, John Aslanides, Amelia Glaese, Nat McAleese, and Geoffrey Irving.
\newblock {Red Teaming Language Models with Language Models}.
\newblock In \emph{EMNLP}, 2022.

\bibitem[Qi et~al.(2019)Qi, Chen, Zhuang, Liu, and Chen]{qi2019network}
Baogui Qi, He~Chen, Yin Zhuang, Shaorong Liu, and Liang Chen.
\newblock A network pruning method for remote sensing image scene classification.
\newblock In \emph{2019 IEEE International Conference on Signal, Information and Data Processing (ICSIDP)}, pp.\  1--4. IEEE, 2019.

\bibitem[Rando et~al.(2022)Rando, Paleka, Lindner, Heim, and Tramèr]{rando2022redteaming}
Javier Rando, Daniel Paleka, David Lindner, Lennart Heim, and Florian Tramèr.
\newblock {Red-Teaming the Stable Diffusion Safety Filter}.
\newblock In \emph{NeurIPS ML Safety Workshop}, 2022.

\bibitem[Robnik-{\v{S}}ikonja \& Kononenko(2008)Robnik-{\v{S}}ikonja and Kononenko]{robnik2008explaining}
Marko Robnik-{\v{S}}ikonja and Igor Kononenko.
\newblock Explaining classifications for individual instances.
\newblock \emph{IEEE Transactions on Knowledge and Data Engineering}, 20\penalty0 (5):\penalty0 589--600, 2008.

\bibitem[Roscher et~al.(2020{\natexlab{a}})Roscher, Bohn, Duarte, and Garcke]{roscher2020explain}
R~Roscher, B~Bohn, MF~Duarte, and J~Garcke.
\newblock Explain it to me--facing remote sensing challenges in the bio-and geosciences with explainable machine learning.
\newblock \emph{ISPRS Annals of the Photogrammetry, Remote Sensing and Spatial Information Sciences}, 3:\penalty0 817--824, 2020{\natexlab{a}}.

\bibitem[Roscher et~al.(2020{\natexlab{b}})Roscher, Bohn, Duarte, and Garcke]{roscher2020explainable}
Ribana Roscher, Bastian Bohn, Marco~F Duarte, and Jochen Garcke.
\newblock Explainable machine learning for scientific insights and discoveries.
\newblock \emph{Ieee Access}, 8:\penalty0 42200--42216, 2020{\natexlab{b}}.

\bibitem[Singh et~al.(2022)Singh, Roy, Setia, and Pateriya]{singh2022estimation}
Harpinder Singh, Ajay Roy, Raj Setia, and Brijendra Pateriya.
\newblock Estimation of chlorophyll, macronutrients and water content in maize from hyperspectral data using machine learning and explainable artificial intelligence techniques.
\newblock \emph{Remote Sensing Letters}, 13\penalty0 (10):\penalty0 969--979, 2022.

\bibitem[Turan et~al.(2023)Turan, Aptoula, Ert{\"u}rk, and Taskin]{turan2023interpreting}
DE~Turan, E~Aptoula, A~Ert{\"u}rk, and G~Taskin.
\newblock Interpreting hyperspectral remote sensing image classification methods via explainable artificial intelligence.
\newblock In \emph{IGARSS 2023-2023 IEEE International Geoscience and Remote Sensing Symposium}, pp.\  5950--5953. IEEE, 2023.

\bibitem[Wijata et~al.(2023)Wijata, Foulon, Bobichon, Vitulli, Celesti, Camarero, Di~Cosimo, Gascon, Long{\'e}p{\'e}, Nieke, et~al.]{wijata2023taking}
Agata~M Wijata, Michel-Fran{\c{c}}ois Foulon, Yves Bobichon, Raffaele Vitulli, Marco Celesti, Roberto Camarero, Gianluigi Di~Cosimo, Ferran Gascon, Nicolas Long{\'e}p{\'e}, Jens Nieke, et~al.
\newblock Taking artificial intelligence into space through objective selection of hyperspectral earth observation applications: To bring the “brain” close to the “eyes” of satellite missions.
\newblock \emph{IEEE Geoscience and Remote Sensing Magazine}, 11\penalty0 (2):\penalty0 10--39, 2023.

\bibitem[Zhou \& Mentch(2023)Zhou and Mentch]{zhou2023trees}
Siyu Zhou and Lucas Mentch.
\newblock Trees, forests, chickens, and eggs: when and why to prune trees in a random forest.
\newblock \emph{Statistical Analysis and Data Mining: The ASA Data Science Journal}, 16\penalty0 (1):\penalty0 45--64, 2023.

\bibitem[Zhou et~al.(2018)Zhou, Zhang, Wang, and Wang]{zhou2018feature}
Yi~Zhou, Rui Zhang, Shixin Wang, and Futao Wang.
\newblock Feature selection method based on high-resolution remote sensing images and the effect of sensitive features on classification accuracy.
\newblock \emph{Sensors}, 18\penalty0 (7):\penalty0 2013, 2018.

\end{thebibliography}
\bibliographystyle{iclr2024_conference}

\clearpage
\appendix
\section{Appendix}

\subsection{Dataset And Model Analysis}\label{sec:data_anal}

\paragraph{Dataset Analysis}
As outlined in Section~\ref{sec:challenge}, the \Hyperview challenge comprises \textbf{1732} training and \textbf{1154} testing patches with a spectral range of \textbf{$462.080$ -- $938.370$ nm} across $150$ bands. The dataset's heterogeneity is evident in varying patch sizes and aspect ratios, challenging the selection of an architectural approach. Notably, approximately $38\%$ of the dataset consists of small patches (up to $32\times32$ pixels), further complicating the analysis. The \Intuition dataset, while spatially identical to \Hyperview challenge, offers a distinct spectral perspective ($192$ bands), providing a broader spectral viewpoint without adding spatial complexity. Analysis of ground truth data reveals a concentrated distribution of soil parameters, with outliers constituting less than $5\%$ of the data. Excluding \textit{pH}, outliers tend to skew right of the mean, while \textit{pH} outliers show more symmetry.

\begin{figure}[ht!]
\centering
\includegraphics[width=0.99\textwidth]{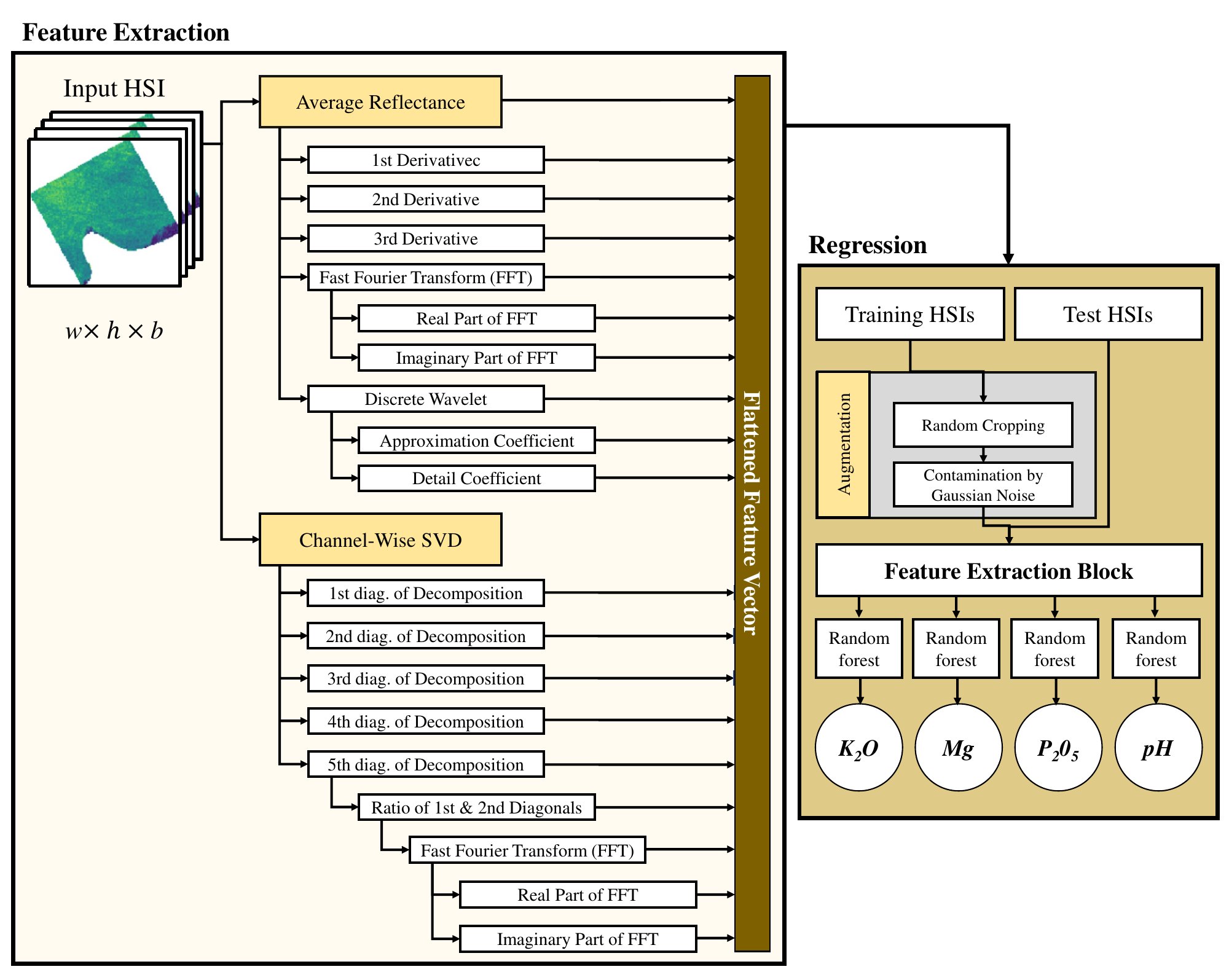}
\caption{A flowchart presenting the \EagleEyes soil parameters' estimation approach. Four regression models (random forests) are trained to retrieve one parameter each, and they operate on manually-designed feature extractors obtained for an input HSI of size $w\times h\times b$, where $w$ and $h$ is its width and height, respectively, and $b$ denotes the number of spectral bands (in this study, $b=150$ or $b=192$). }
\label{fig:eagleeyes}
\end{figure}

\begin{figure}[ht]
\centering
\includegraphics[width=1.0\textwidth]{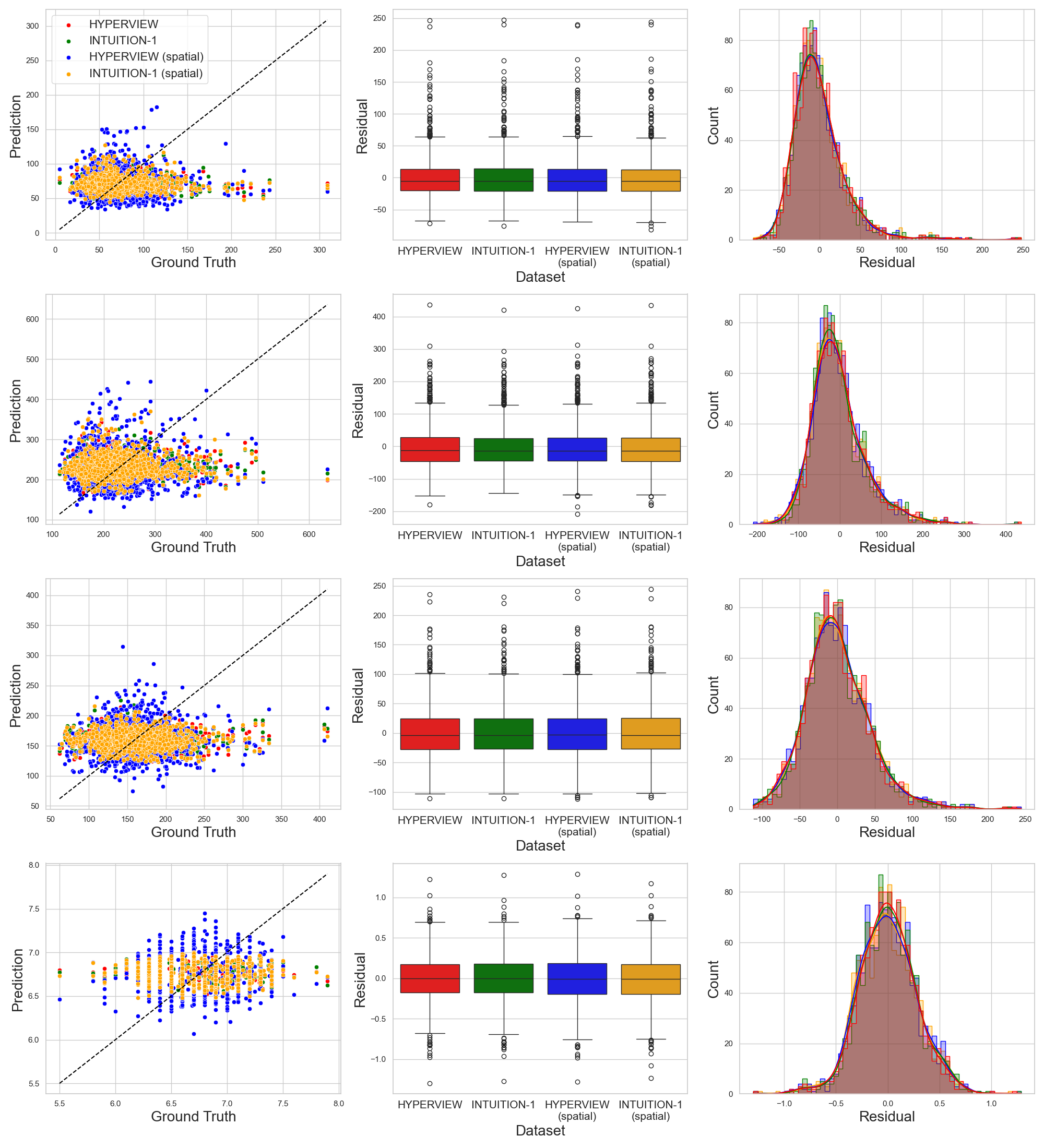}
\caption{Residuals visualization for each soil parameter across different model versions. Each row represents a soil parameter, in the order of \textit{phosphorus}, \textit{potassium}, \textit{magnesium}, and \textit{pH}. Within each row: the left panel shows a scatter plot (ground truth on the $x$-axis, predicted values on the $y$-axis), the middle panel displays a boxplot of residuals, and the right panel presents a histogram of the residuals.}
\label{fig:residuals_all}
\end{figure}

\paragraph{Model Analysis}
The superior \EagleEyes model from the \Hyperview challenge builds upon classic ML models with the combination of feature engineering for supervised regression tasks (Figure~\ref{fig:eagleeyes}). This model was developed in four variations, with and without spatial feature extraction, tailored for HSIs of 150 (\Hyperview dataset) and 192 (\Intuition dataset) spectral bands. We specify the inclusion of spatial features by adding ``spatial" in parentheses when referencing a model (e.g., \Hyperview (spatial) benefits from spatial features across 150 bands, while \Intuition focuses solely on spectral features over 192 bands). Performance evaluation involved juxtaposing predicted values against actual ground truth data, and analyzing residuals to assess accuracy. Figure~\ref{fig:residuals_all} showcases the \EagleEyes model's iterations across soil parameters. Despite uniform performance across parameters, the model's accuracy was far from ideal. Residual analysis, through boxplots and histograms, indicated a Gaussian distribution with a mean of $0$, but notable variance points to significant deviations for many samples. Scatter plots in Figure~\ref{fig:residuals_all} reflect this, showing predictions primarily near the zero-error line but with widespread variation. This distribution leads to higher residuals, especially for values outside the model's learned prediction range, highlighting challenges with distinct or outlier data.

\subsection{Mathematics Behind SHAP}\label{sec:math}

The Shapley value $\phi_i(x)$ for feature $i$ and observation $x$ are defined as 
$$
\varphi_{i}(x)=\sum_{S\subseteq P\setminus \{i\}}{\frac {|S|!\;(p-|S|-1)!}{p!}}(v_{S\cup \{i\}}(x)-v_S(x)),
$$
where $P$ is a set of all features, $p=|P|$ is the number of features, $v_S(x)$ is the prediction calculated for model $f$ after features from set $S$ are set, typically it is the mean prediction for the conditional distribution with fixed values of the features in the set $S$, i.e.,~$v_S(x) = E_{x|S} f(x)$, but there are also other alternatives used if features are correlated~\citep{covert2020understanding}. Shapley values are calculated locally for a specific feature and observation, but they can also be aggregated to get the global importance of a selected variable  $imp(i) = \sum_{j=1}^N |\varphi_i(x_j)|$,
or attribution of a group of features ($F \subseteq P$)\;$\varphi_F(x) = \sum_{i\in F} \varphi_i(x)$.
In our case, we make use of this ability to aggregate shapley values according to hyperspectral bands or specific transformation methods.

\subsection{SHAP in Residuals}

In our extended analysis, we closely examined instances of significant overestimation and underestimation, as well as the “best” prediction outcomes, across all model variants. We consistently observed the same set of features appearing in each scenario, although their importance rankings varied. This pattern makes it challenging to identify particular features responsible for overestimation and underestimation directly. Moreover, investigating specific prediction cases did not yield insights that could be generalized to understand the model's overall behavior or the characteristics of the underlying data more broadly.

\newpage
\subsection{SHAP Visualizations}\label{sec:shap_vis}

\begin{figure}[ht]
\centering
\includegraphics[width=1.0\textwidth]{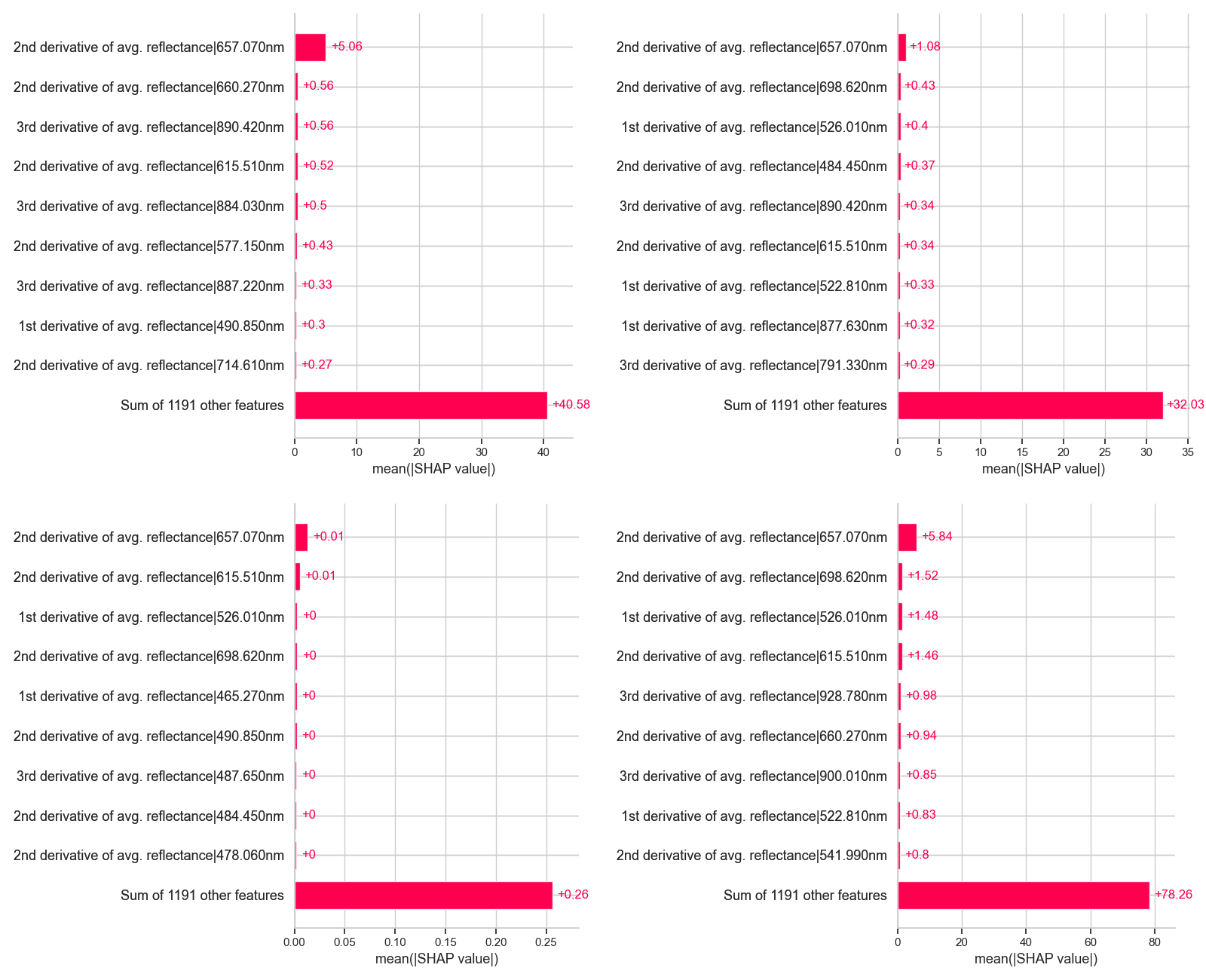}
\caption{Visualization of the top $10$ features identified by Shapley values, ordered by their contribution to model predictions in the \EagleEyes model trained for the \Hyperview challenge. The panels show, clockwise from top left, \textit{phosphorus}, \textit{potassium}, \textit{magnesium}, and \textit{pH}.}
\label{fig:shap_all_soil}
\end{figure}

\begin{figure}[ht]
\centering
\includegraphics[width=1.0\textwidth]{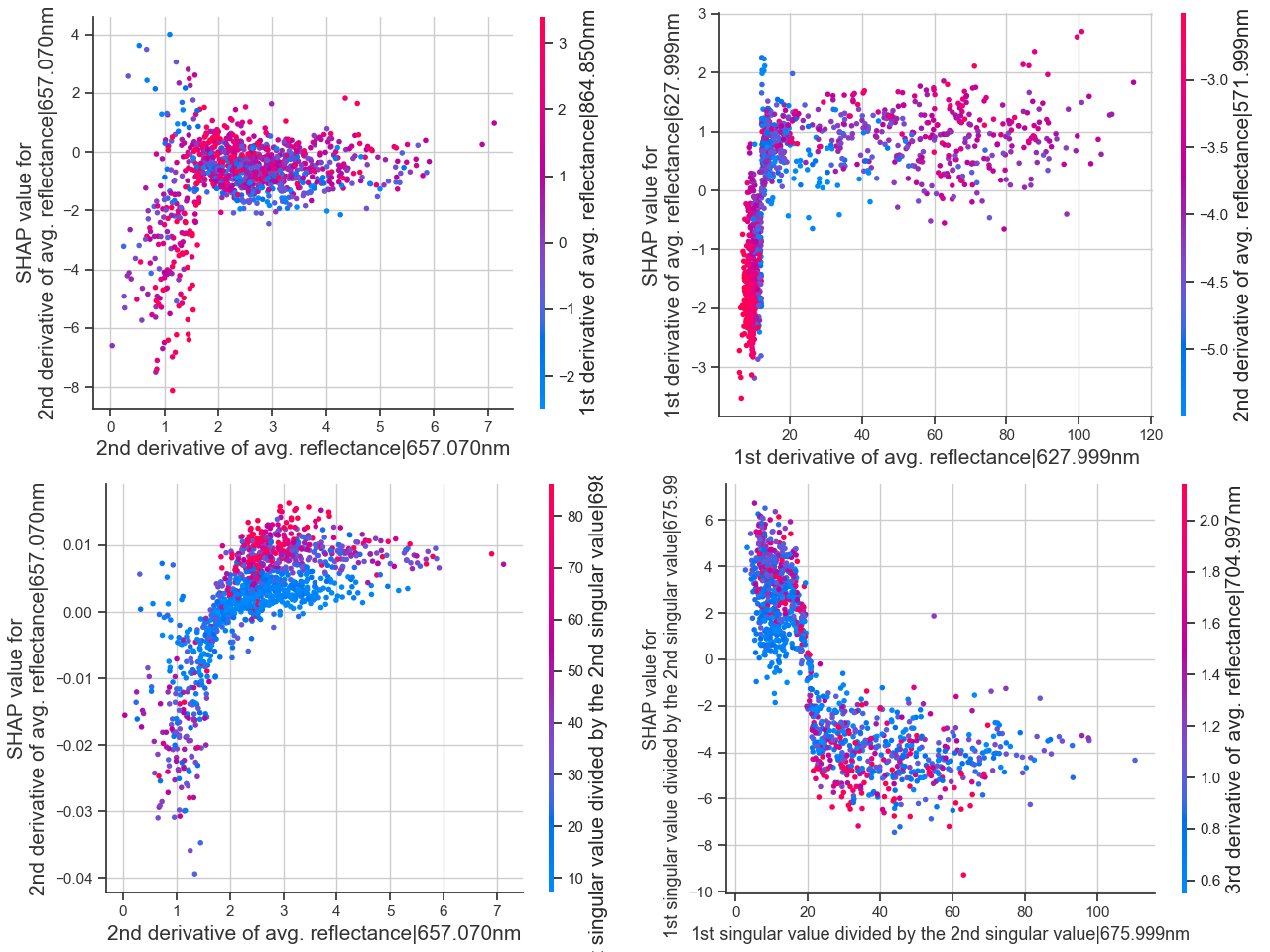}
\caption{SHAP dependency plots showcasing the most influential feature in each iteration of the \EagleEyes model. The arrangement is as follows: \Hyperview (top left), \Intuition (top right), \Hyperview (spatial) (bottom left), and \Intuition (spatial) (bottom right). The $x$-axis represents the variation in feature value, while the $y$-axis depicts the corresponding change in Shapley values.}
\label{fig:shap_dependency_all}
\end{figure}

\begin{figure}[ht]
\centering
\includegraphics[width=1.0\textwidth]{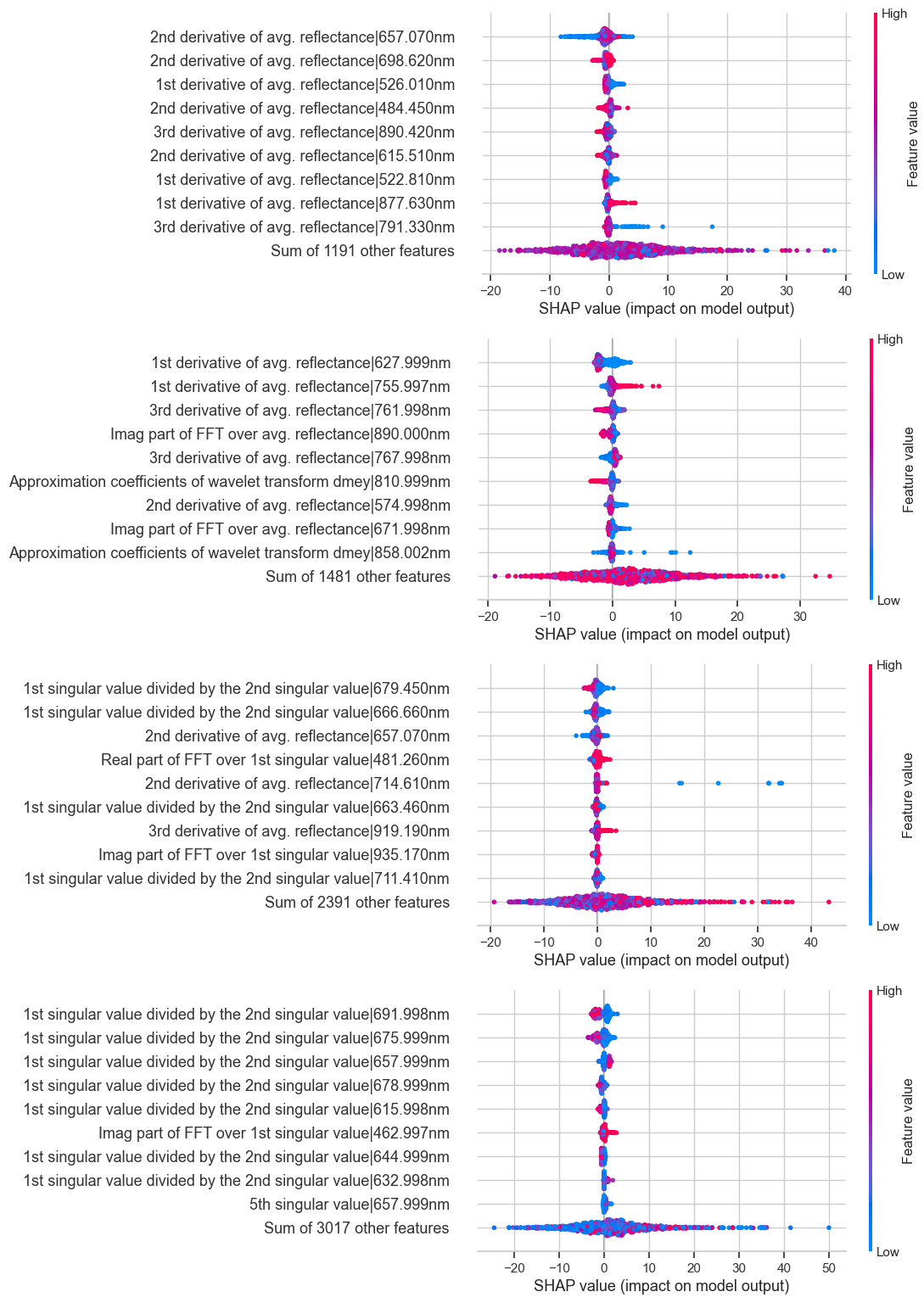}
\caption{SHAP beeswarm plots for each iteration of the \EagleEyes model, starting from the top \Hyperview, \Intuition, \Hyperview (spatial), and \Intuition (spatial), displaying how feature values contribute to predictions. Features are displayed in descending order of influence.}
\label{fig:shap_all_models}
\end{figure}

\newpage
\subsection{SHAP Aggregation Visualizations}\label{sec:aggr_vis}

\begin{figure}[ht]
\centering
\includegraphics[width=1.0\textwidth]{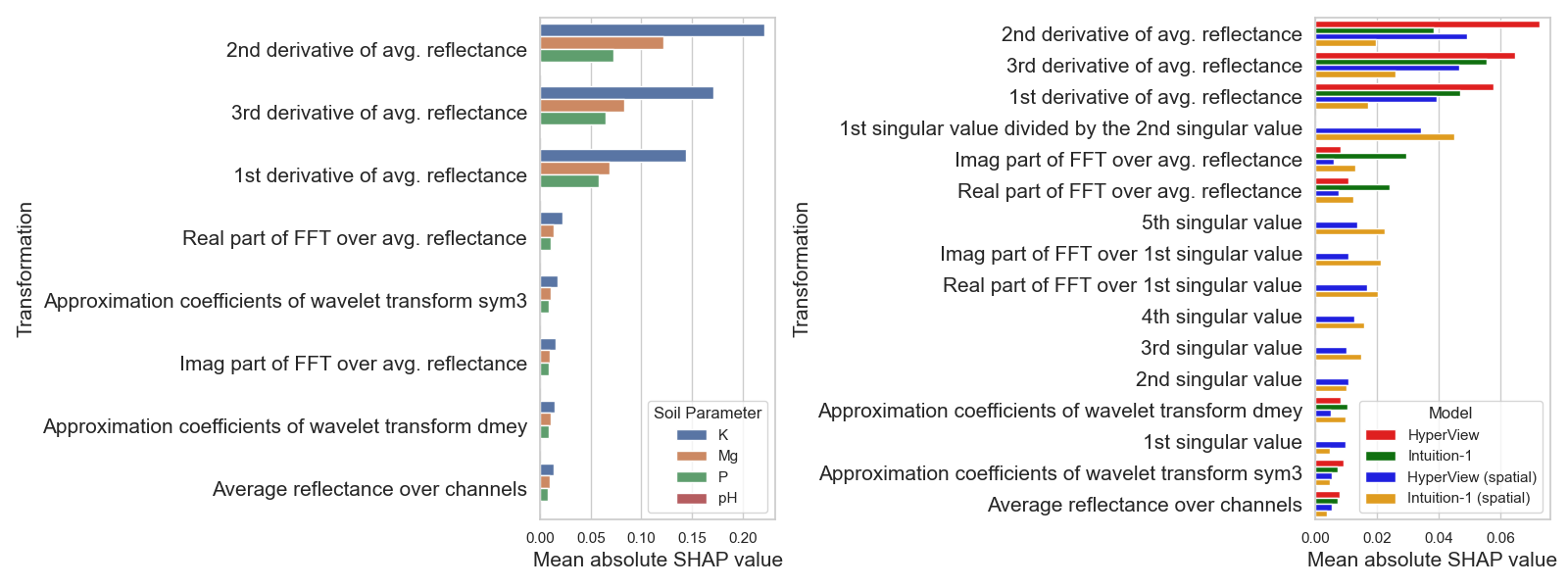}
\caption{Shapley values for transformation aggregation in the \EagleEyes models, showing feature importance across soil parameters (left) and model iterations (right).}
\label{fig:shap_transformation}
\end{figure}

\begin{figure}[ht]
\centering
\includegraphics[width=1.0\textwidth]{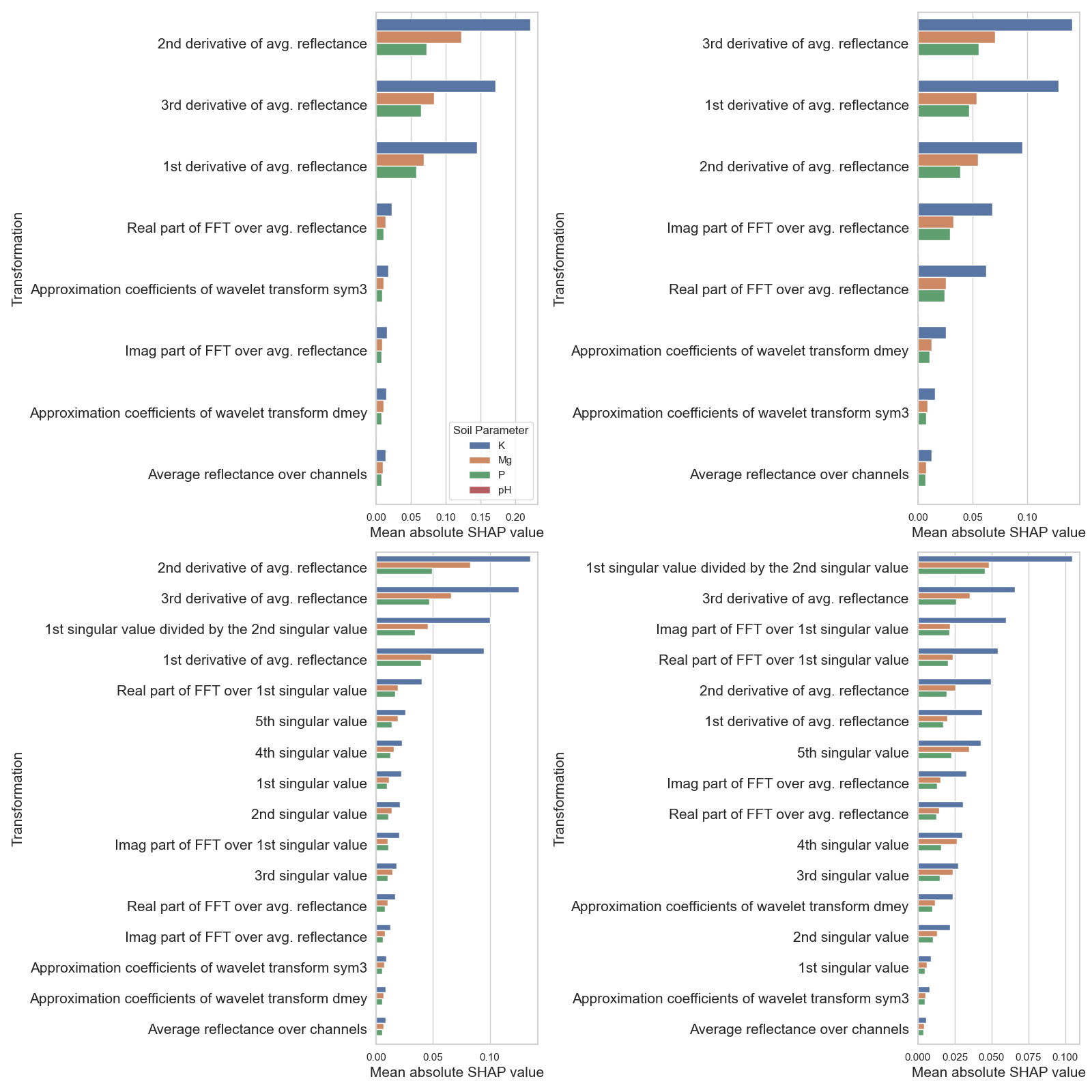}
\caption{Shapley values for transformation aggregation across each iteration of the \EagleEyes model, presented as follows: \Hyperview (top left), \Intuition (top right), \Hyperview (spatial) (bottom left), and \Intuition (spatial) (bottom right), illustrating feature importance across various soil parameters.}
\label{fig:shap_transformation_soil}
\end{figure}

\end{document}